\documentclass[runningheads]{llncs}
\usepackage{graphicx}
\usepackage{caption}
\usepackage{subcaption}
\usepackage{xcolor}
\pdfoutput=1 
\begin{document}
\title{How Well Can You Articulate that Idea? Insights from Automated Formative Assessment}
\titlerunning{How Well Can you Articulate that Idea?}
% \author{Anonymous}
% \authorrunning{Anon.}
\author{Mahsa Sheikhi Karizaki\inst{1} \and
Dana Gnesdilow\inst{2}
\and
Sadhana Puntambekar\inst{2}
\and
Rebecca J. Passonneau\inst{1}\orcidID{0000-0001-8626-811X}}
\authorrunning{M. Sheikhi et al.}
% \institute{Anonymous Institution}
\institute{Pennsylvania State University, State College, PA 16801, USA \\
\email{(mfs6614, rjp49)@psu.edu}
\and
University of Wisconsin-Madison, Madison, WI, USA\\
\email{gnesdilow@wisc.edu, puntambekar@education.wisc.edu}}

\maketitle              % typeset the header of the contribution
\begin{abstract}
Automated methods are becoming increasingly integrated into studies of formative feedback on students' science explanation writing. Most of this work, however, addresses students' responses to short answer questions. We investigate automated feedback on students' science explanation essays, where students must articulate multiple ideas. Feedback is based on a rubric that identifies the main ideas students are prompted to include in explanatory essays about the physics of energy and mass, given their experiments with a simulated roller coaster. We have found that students generally improve on revised versions of their essays. Here, however, we focus on two factors that affect the accuracy of the automated feedback. First, we find that the main ideas in the rubric differ with respect to how much freedom they afford in explanations of the idea, thus explanation of a natural law is relatively constrained. Students have more freedom in how they explain complex relations they observe in their roller coasters, such as transfer of different forms of energy. Second, by tracing the automated decision process, we can diagnose when a student's statement lacks sufficient clarity for the automated tool to associate it more strongly with one of the main ideas above all others. This in turn provides an opportunity for teachers and peers to help students reflect on how to state their ideas more clearly.

\end{abstract}
\section{Introduction}
\label{sec:intro}

Science writing has been found to support science learning, especially to enhance students' inquiry and reasoning skills~\cite{gereEtAl_writing&learning_2019,grahamEtAl_meta-analysis_rev-ed-res20,huerta&garza_litrev-science-writ_edpsychrev19,klein_boscolo_2016}.  Artificial Intelligence incorporated into formative assessment of science writing provides many opportunities to support that learning, such as individual feedback for revision support~\cite{gerard&linn_revis-science-explanation_comp&ed22,leeEtAl_automatic-feedback_scied19,zhangEtAl_erevise_iaai19}. However, using AI tools for science essay revision feedback is still novel, therefore little is known about how accurately AI tools can provide feedback on student writing, much less factors that affect performance. In a project that provides a web-delivered short curriculum on middle school roller coaster physics, students are prompted to write essays explaining the main ideas in the curriculum, then revise their essays based on automated feedback. We find that the accuracy of automated detection of ideas in students' essays depends not only on how clearly the students express themselves, but also on the inherent distinctiveness of propositions that express the main ideas.

The web-delivered curriculum uses an existing automated tool for detecting main ideas in short passages written to the same prompt, where the feedback to students is in the form of a checklist. The tool supports substitution of alternative methods to convert strings of words to vector representations. To optimize accuracy of the feedback, we tested multiple ways to generate semantic vectors on a set of manually labelled student data, before classroom deployment. After classroom use, we manually labeled a new set of essays. We were thus able to confirm the accuracy on the new classroom sample. More interestingly, we were able to assess factors that affect accuracy that also correlate with characteristics of middle school writing. The main measure used by the tool to determine whether an essay contains a given idea is to compare the cosine similarity of vectors of student clauses to vectors of main idea clauses in a pre-defined model.  Analysis of the distributions of pairwise cosine similarities of student phrases to model ideas reveals patterns that can be used to explain lower performance of the tool. For example, some essays have multiple phrases that have moderately good cosine similarity to multiple model ideas.  We also observe that certain ideas, such as the law of conservation of energy, are more accurately detected, which we attribute to the formulaic expressions used in statements of the law.

The first half of the paper presents related work, the curriculum, our datasets, and how we chose a vectorization method for the essay feedback software.  Sections \ref{sec:distinctiveness-of-ideas}-\ref{sec:discussion} present our analyses of the software accuracy, and how investigations of distributions of cosine similarities reveal differences in distinctiveness of main ideas, as well as differences in clarity of students' statements of these ideas.

\section{Related Work}
\label{sec:related}

Automated methods to support formative feedback on students' responses to open-ended questions in STEM subjects often focus on helping students 
revise their answers. 
Formative feedback, meaning feedback during a unit or course to support further learning, has been found to be most beneficial when it focuses on the \textit{what, how and why} of a problem rather than on verification of results~\cite{shute_formative-feedback_edres08}. A series of papers from Linn and her students have investigated the use of automated guidance in support of short answer explanations from middle school students. They have compared automated feedback alone and in combination with information about the personalized nature of the feedback~\cite{tansomboonEtAl_automated-revision_ijaied17}, alone or in combination with students providing feedback on a sample essay~\cite{gerardEtAl_essayrevision_icls16}, and finally alone or in combination with an interface that models the revision process \cite{gerard&linn_revis-science-explanation_comp&ed22}. In all three cases, automated guidance was from the C-rater-ML tool~\cite{heilman&madnani_craterml_semeval13}, reported to have a 0.72 Pearson correlation with human guidance in~\cite{tansomboonEtAl_automated-revision_ijaied17}. The overarching aim of these studies was to understand how to improve students' use of evidence in science explanation, given that students often lack revision skills. In sum, a combination of automated feedback with explicit modeling of revision had the greatest benefit.

A similar theme was investigated by the Concord Consortium \cite{leeEtAl_auto-feedb-revis_sciedtec21,zhuEtAl_auto-feedback-revision-sci-arg_comp&ed20,leeEtAl_automatic-feedback_scied19}, mostly with high school students, aimed at improving students' understanding of uncertainty in science~\cite{pallantEtAl_second-student-sci-arg_geoed20}. All studies relied on C-rater-ML, achieving QWK scores with humans between 0.78 and 0.93, depending on the study. The first study found that students overwhelmingly improved their revisions given the automated feedback \cite{leeEtAl_automatic-feedback_scied19}. The second study compared generic argumentation feedback to student-specific feedback through use of C-rater-ML, with the latter leading to greater improvements in revisions~\cite{zhuEtAl_auto-feedback-revision-sci-arg_comp&ed20}. The third study compared feedback on argumentation writing alone or in combination with feedback on students' use of the science simulations and resulting data~\cite{leeEtAl_auto-feedb-revis_sciedtec21}. Revisions from students in both conditions improved over the original responses; only those who also received feedback on the simulations improved their use of data.

In contrast to the work mentioned above on short answer responses, our work provides formative feedback on science explanation essays, which ask students to explain multiple ideas. There is relatively little work on automated formative feedback for essay revision. Zhang et al. \cite{zhangEtAl_erevise_iaai19} presented eRevise, which provides rubric-based feedback on students' use of evidence for source-based essays. By comparing multiple approaches, the authors found that reliance on word embeddings had the best combination of performance accuracy and ability to provide student-specific feedback. Tests with middle school students showed that eRevise led to improved scores on revisions. Our work relies on PyrEval~\cite{gao-etal-2019-automated}, a publicly available software package, that also uses word embeddings. It can be used with a rubric that lists the main ideas a student explanation should include~\cite{singh_automated_2022}. In separate work, we report that PyrEval feedback helps students improve in their revised science essays. As described below, the tool has high accuracy on identifying when ideas are present or missing. We find, however, that accuracy across main ideas varies, depending on the distinctiveness of an idea from others in the curriculum, and on the clarity of students' sentences. From discussions with teachers and students about why PyrEval marked an idea as missing that students thought they had stated, we conclude that even somewhat inaccurate feedback can help students reflect on the way they articulate their ideas.

\section{Roller Coaster Physics Curriculum}
\label{sec:curric}

\begin{figure}[t!]
    \centering
    \small
    \begin{subfigure}[t]{0.33\textwidth}
    \begin{tabular}{c | r | l}
    \multicolumn{1}{c|}{ID} & 
        \multicolumn{1}{c|}{Sim} &      
            \multicolumn{1}{c}{Text Description} \\\hline
    1    & 0.69 &  The greater the height, the \\
         &      &  greater the potential energy (PE) \\\hline
    2    & 0.77 &  As the cart moves downhill, PE de- \\
         &      &  creases and kinetic energy increases\\\hline
    3    & 0.60 &  The total energy of the system is \\
         &      &  always the sum of PE and KE\\\hline
    4    & 0.75 &  The law of conservation of energy \\
         &      &  states that energy cannot be created \\
         &      &  or destroyed, only transformed\\\hline
    5    & 0.75 &  The initial drop should be \\
         &      &  higher than the hill \\\hline
    6    & 0.70 &  Higher mass of the cart corresponds \\
         &      &  to greater total energy of the system
    \end{tabular}
    \caption{The Six Main Ideas.}
        \label{fig:main-ideas}
    \end{subfigure}
    % \begin{subfigure}[t]{0.4\textwidth}
    % \begin{tabular}{c|l}
    % \multicolumn{1}{c|}{ID} & \multicolumn{1}{c}{Text Description} \\\hline
    % 1    &  The greater the height, the \\
    %      &  greater the potential energy (PE) \\\hline
    % 2    &  As the cart moves downhill, PE de- \\
    %      &  creases and kinetic energy increases\\\hline
    % 3    &  The total energy of the system is \\
    %      &  always the sum of PE and KE\\\hline
    % 4    &  The law of conservation of energy \\
    %      &  states that energy cannot be created \\
    %      &  or destroyed, only transformed\\\hline
    % 5    &  The initial drop should be \\
    %      &  higher than the hill \\\hline
    % 6    &  Higher mass of the cart corresponds \\
    %      &  to greater total energy of the system
    % \end{tabular}
    % \caption{The Six Main Ideas}
    %     \label{fig:main-ideas}
    % \end{subfigure}
    \hfill
    \vspace{.2in}
    \begin{subfigure}[t]{0.44\textwidth}
    \raisebox{-.1in}{\includegraphics[width=\textwidth]{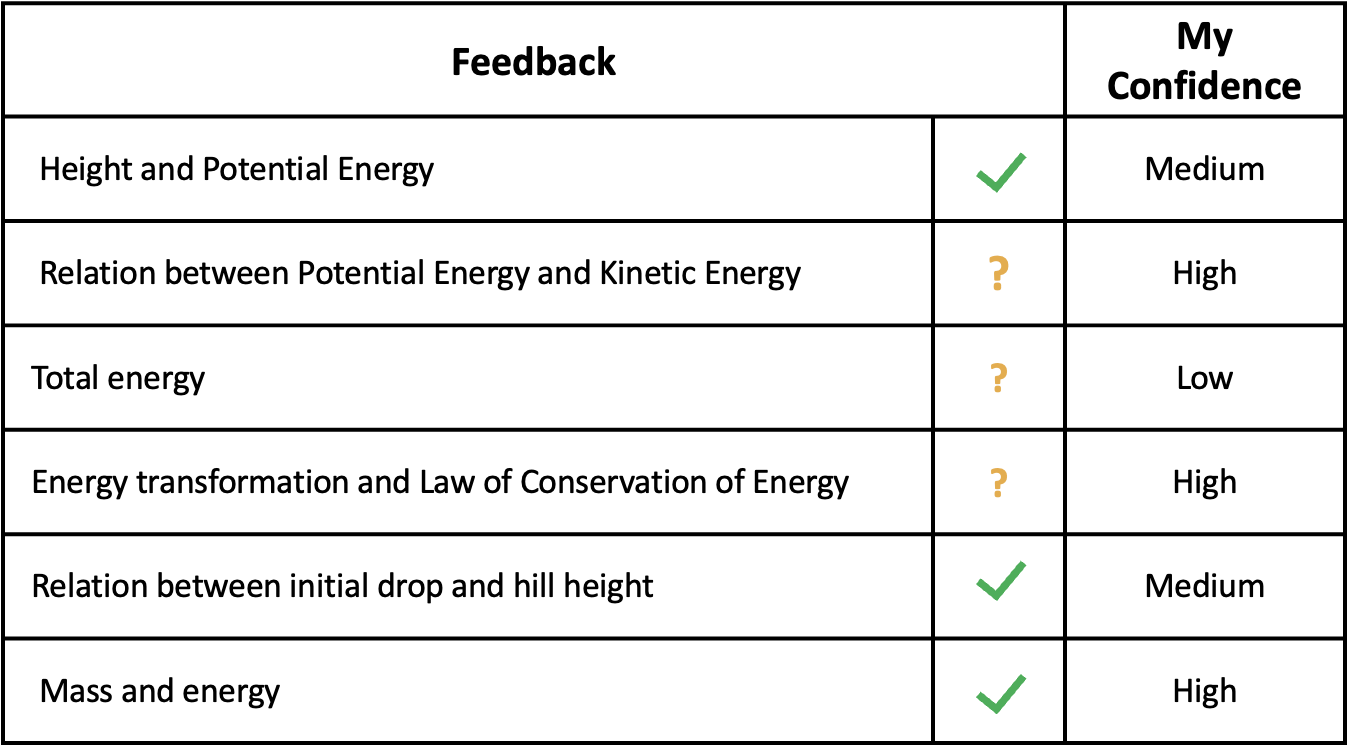}}
        \caption{Sample Feedback Checklist: a green check mark means PyrEval detected the idea; a gold question mark indicates PyrEval did not. The 'My Confidence' column reflects PyrEval's average accuracy for a given idea.}
        \label{fig:feedback}
    \end{subfigure}
    \vspace{-.3in}
    \caption{Essay 1 Main Ideas with average cosine similarities within the pyramid content unit (Sim), and sample feedback checklist.}
    \label{fig:main-ideas-and-prompts}
\end{figure}

During a 2-3 week design-based physics unit, middle school students learned about the physics of energy and energy transfer to design a safe and fun roller coaster. The curriculum was delivered through a digital notebook in which they conducted virtual experiments using a roller coaster simulation, recorded their data, and answered multiple choice and open-ended questions before submitting explanation essays and essay revisions. For each of two essays, an essay prompt guided them about which ideas to include, such as the influence of height on potential energy, or how energy transfers as the roller coaster car moves down the initial drop. Here we focus on the first essay which addressed the six ideas shown in Fig.~\ref{fig:main-ideas}. In separate work, we report that students' revised essays improved based on the automated feedback (Anon), which was presented to students in the form of a checklist, as shown in Fig~\ref{fig:feedback}.

\section{Data} 
\label{sec:data}

Three datasets are utilized in this work. We first describe a dataset of 39 essays that were annotated for presence of the six main ideas shown in Fig~\ref{fig:main-ideas}. These were written by students in year 2 of the 4-year project, and used to select the best semantic vector method and hyperparameters for PyrEval to use in the remaining two years of the project. We refer to this dataset as Ground Truth 1 (GT1). We then describe a set of 60 essays and their revised versions written in year 3 of the project, where we verified the accuracy of the feedback PyrEval provided to students; we refer to this dataset as Ground Truth 2 (GT2). Finally, we use a historical dataset we refer to as MidPhys that has middle school students' written responses to physics questions on a broad set of topics besides energy. We test its utility for fine-tuning the vector methods used by PyrEval.

In year 2 of the project, we selected 7 high quality student essays to construct a new PyrEval model, referred to as a pyramid, and 39 additional essays of varying quality to create Ground Truth 1 (GT1).  Three annotators from the project worked independently to label the 39 essays for presence of each main idea. We then arrived at a consensus labeling, which was updated several times while testing alternative pyramids. PyrEval can automatically generate a pyramid from five reference essays using an algorithm that groups statement of the same idea expressed in different reference essays into content units.  Each content unit then has a weight representing how many of the reference essays stated the idea. To support automated feedback on the six main ideas in Fig.~\ref{fig:main-ideas-and-prompts}, we aimed for a pyramid with exactly six content units of weight 5 (the maximum weight) corresponding one-to-one to the six main ideas in the curriculum. From the 7 high quality essays we generated 21 distinct pyramids based on all subsets of 5.  After selecting the pyramid with the best performance on GT1, we manually edited the 5 corresponding reference essays to further improve the final pyramid.

From essays written in year 3 of the project, essays and essay revisions from 60 students for essay 1 were coded to create the 120 essays in Ground Truth 2 (GT2). Raters examined the feedback provided by PyrEval, which indicates presence or absence of each of the six main ideas, and labeled the feedback as correct or incorrect. Interrater reliability was measured on 20\% of the essays from two researchers working independently. Substantial agreement of Cohen's Kappa = 0.768 was achieved. Discrepancies between raters were discussed, then one of the researchers coded the remainder of the data.

The MidPhys dataset consists of 11,245 constructed responses from middle school students to 55 physics questions, collected from a decade of studies. We tested whether combining this data with a corpus used by PyrEval could improve the tool's ability to handle middle school writing about physics. 

\section{Selection of a Semantic Vector Method for PyrEval}
\label{sec:vector-method}

\begin{table}[t!]
\small
    \centering
    \begin{tabular}{l|r|r|r}
     \multicolumn{1}{c|}{Method}    &  
        \multicolumn{1}{c|}{\textit{topk}}   &  
            \multicolumn{1}{c|}{$t$}   & 
                \multicolumn{1}{c}{Accuracy}   \\\hline
    WTMF     &  3 & 0.55 & 0.795 \\
    WTMF with MidPhys     &  3 & -0.01 & 0.675\\
    WTMF Refinement     &  3 &  -0.01 & 0.705 \\
    BERT     &  3 & 0.85 &  0.752\\
    Fine Tuned BERT & 3 & 0.83 & 0.752 \\
    BERT + WTMF     & 3 & 0.83 & 0.756 \\         
    \end{tabular}
    \caption{Comparison of Six Semantic Vector Methods on GT1}
    \label{tab:vectormethods}
\end{table}

PyrEval is a publicly accessible off-the-shelf tool that has been used in previous work on student essays \cite{singh_automated_2022}. We used PyrEval because it requires little training data compared to other methods, and it is highly modular. For example, it supports user selection of different semantic vector methods. It also has a set of hyperparameters to adjust PyrEval to a new dataset, using grid search~\cite{singh_automated_2022}. 

PyrEval has been described elsewhere~\cite{gao-etal-2019-automated,singh_automated_2022}. Here we briefly outline the preprocessing steps, the pyramid, and the assessment method. Step 1 of the preprocessing converts essay sentences into clauses, and step 2 converts each clause into a semantic vector. From a sample of five high quality exemplar essays, PyrEval can automatically construct a content model known as a pyramid, where each content unit represents different ways of expressing the same content. An algorithm groups clause vectors from the five exemplar essays into content units (CUs) of at most five vectors, each from a different input exemplar essay, based on maximizing the average of all pairwise cosine similarities of vectors within a single CU. CUs with fewer than five vectors have lower importance: pyramids have a long tail with increasingly many CUs for each  lower weight. 
%As described in the preceding section, we created a pyramid with a one-to-one mapping of the main ideas students are prompted to explain (cf. Fig.~\ref{fig:main-ideas-and-prompts}), and the weight five content units. 
The \textit{Sim} column of Fig.\ref{fig:main-ideas} shows the average pairwise cosine similarity of the five vectors within each of the six main idea content units.

A second algorithm matches clause vectors from a student essay to the pyramid content units, using average cosine similarity of a student vector to all the vectors in a content unit as the criterion. The matching is done by a variant of a greedy algorithm for maximal independent set (MIS) selection, where the input to the algorithm is a hypergraph. Each node in the hypergraph is a set of possible matches between a content unit in the pyramid model, and the most similar clauses in a student essay, given a user determined similarity threshold~\cite{singh_automated_2022}. The output of the MIS algorithm is the set of nodes that represent the best matches of phrases from a student essay to the pyramid model. We performed grid search over two key parameters: \textit{topk} for the number of candidate matches to each pyramid node, and $t$, the threshold cosine similarity value of a student phrase to a pyramid content unit to be added to the assessment hypergraph.

PyrEval's github includes a matrix factorization vector method, WTMF~\cite{guo-diab-2012-modeling}, and a high quality WTMF training corpus. The output of WTMF is a vector dictionary for words, and a method to construct vectors for unseen sentences and phrases from the fixed vector dictionary. In contrast to WTMF, BERT~\cite{devlin-etal-2019-bert} is a high-quality pretrained language model that can be used to map input words and phrases to contextualized vectors, rather than using a fixed vector dictionary. We compared six semantic vector methods for use in PyrEval: 1) WTMF with its original corpus;  2) WTMF on the original corpus combined with MidPhys (see above); 3) refinement of the WTMF vector space; 4) BERT; 5) BERT fine-tuned on MidPhys; 6) fine-tuned BERT vectors concatenated with WTMF vectors.

The method for refining word embeddings presented in \cite{yu-etal-2017-refining} adjusts sentiment words with opposite polarity to have lower cosine similarity. It relies on a human ranking of sentiment words.  Our goal was to adjust the cosine similarities of physics vocabulary to be more distant for word pairs like ``potential'' and ``kinetic,'' and to do so using an automated ranking rather than relying on manual labels. We used tf-idf scores computed from the MidPhys corpus to rank words. %, treating each student answer as a distinct document, and using the term frequency of words in the entire corpus. %After that, the refined word embeddings were incorporated into PyrEval to assess the newly adjusted embeddings on the model's performance.

Table~\ref{tab:vectormethods} presents the comparison of the six semantic vector methods on the GT1 dataset. Along with accuracy on GT1, we show the values of the two hyperparameters \textit{topk}, and $t$. As shown, none of the methods improved over WTMF. The results in the remainder of the paper thus use this vector method. Combining the MidPhys corpus with the WTMF corpus, which consists largely of definitional sentences, lowered the accuracy; the refinement method also lowered the accuracy, but not quite as much. The three methods with BERT had similar accuracies to one another that were still not as high as WTMF.

\section{Feedback Accuracy and Distinctiveness of Ideas}
\label{sec:distinctiveness-of-ideas}

%The feedback students receive is in the form of a checklist where each main idea prompt is given a check if PyrEval detected the idea in the essay, and a question mark if PyrEval did not, along with an indicator of how consistently correct PyrEval is on a given idea, shown as \textit{high, medium, low} (cf. Fig.~\ref{fig:feedback}). 
Table~\ref{tab:pyreval-acc} shows that accuracy on the GT2 dataset is somewhat lower than on GT1, but still quite good. Lower accuracy on GT2 is to be expected, given that GT1 was relatively small in size, and we tuned PyrEval to have high accuracy on GT1. Accuracies are broken down into accuracy at detecting the presence of an idea (positive accuracy), versus detecting the idea is not present (negative accuracy). Use of MIS for matching student ideas to the pyramid model (see above) translates to the goal of matching as many possible essay ideas to content units in a manner that achieves the highest average cosine similarity per matched idea. This approach inherently favors positive accuracy over negative accuracy.

\begin{table}[t]
    \centering
    \setlength{\tabcolsep}{15pt}
    \begin{tabular}{l |r|r|r}
    \multicolumn{1}{c|}{Dataset}    &
        \multicolumn{1}{c|}{Pos. Acc.}    &
            \multicolumn{1}{c|}{Neg. Acc.}   &  
                \multicolumn{1}{c}{Acc.}    \\\hline
    GT1    &  80.64  &  76.56  & 79.50 \\
    GT2 O  &  73.73  &  77.14  & 74.72 \\
    GT2 R  &  77.00  &  55.32  & 74.17 \\
    GT 2   &  75.53  &  70.39  & 74.44 \\
    All    &  76.78  &  70.05  & 75.47
    \end{tabular}
    \caption{PyrEval positive, negative and total accuracies (as percentages).}
    \label{tab:pyreval-acc}
\end{table}

\begin{table}[b!]
    \centering
    \setlength{\tabcolsep}{10pt}
    \begin{tabular}{l|r|r|r|r|r|r}
        \multicolumn{1}{c|}{Dataset} &
            \multicolumn{1}{c|}{MI 1} &
                \multicolumn{1}{c|}{MI 2} &  
                    \multicolumn{1}{c|}{MI 3} &
                        \multicolumn{1}{c|}{MI 4} &
                            \multicolumn{1}{c|}{MI 5} &
                                \multicolumn{1}{c}{MI 6} \\\hline
    GT1    & 76.92  & 82.05  & 69.23  & 89.74  & 71.79  & 84.62  \\
    GT2 O  & 63.33  & 56.66  & 66.66  & 91.66  & 86.66  & 83.33  \\
    GT2 R  & 63.33  & 61.66  & 76.66  & 86.66  & 86.66  & 70.00  \\
    GT2    & 63.33  & 59.16  & 71.66  & 89.16  & 86.66  & 76.66  \\
    All    & 66.66  & 64.77  & 71.06  & 89.30  & 83.01  & 78.61  \\
    \end{tabular}
    \caption{Accuracy on Main Ideas 1-6 (as percentages).}
    \label{tab:main-idea-acc}
\end{table}

Accuracy across the six ideas in the curriculum is not uniform, as illustrated in Table~\ref{tab:main-idea-acc}. The accuracy annotation in the feedback checklist shown in Fig.~\ref{fig:feedback} is based on GT1 results. However, the relative accuracy per idea is different on GT1 versus the overall set of 159 essays. In all (last row of Table~\ref{tab:main-idea-acc}), the three ideas PyrEval identifies most accurately ($\geq 78\%$) are, in descending order: statements that define the law of conservation of energy (main idea 4; LCE), the statement that the roller coaster initial drop must be higher than the hill that follows (main idea 5), and the statement that greater mass of the cart results in greater total energy (main idea 6). Main idea three has moderate accuracy (71.06\%). Main ideas one and two have much lower accuracy of 66.66\% and 64.77\%, respectively.   %Different pyramid models we tested (cf. section~\ref{sec:data}) had different overall accuracy, as well as somewhat different accuracies on the main ideas, but the trend was always similar to what we see in Table~\ref{tab:main-idea-acc}. 

We attribute the particularly high accuracies for the fourth and fifth main ideas to two factors. One key factor is that certain explanatory or definitional statements have a very narrow range of paraphrases.  While the word order to state the LCE can vary, and some of the important relations can be expressed with different terms (e.g., \textit{transformed, transferred}), there is a relatively canonical formulation of the LCE.  Similarly, for the idea that the roller coaster cart must have a high initial drop to have enough energy to get over a following hill, the range of possible paraphrases is rather narrow.  This is related to the second factor, which is that the concepts in the fourth and fifth main ideas are consistently expressed using vocabulary terms that do not get used to state the other ideas: \textit{law, conservation, transformed, created, destroyed, transferred} for main idea 4; \textit{initial, drop} for main idea 5.  In contrast, the terms \textit{potential energy, kinetic energy, increase, decrease, total energy} are used in statements of all other main ideas. We note that while the term \textit{mass} is unique to idea 6, the embedding for \textit{mass} is fairly close to the embeddings for \textit{energy, potential} and \textit{kinetic}. Further, students explain the relation between PE and energy, or between PE and KE (main ideas one and two) in a large variety of ways. 

Cosine similarity as a measure of the distinctiveness of pairs of main ideas is shown in the last column of Table~\ref{tab:confusable-main-ideas}. It gives the average cosine similarities of all pairs of vectors from two distinct main ideas, for each pair of main ideas. Averaging across pairs gives the lowest averages for main ideas 4 (0.271) and 5 (0.228), moderate for 3 (0.371), and between 0.427 and 0.431 for 6, 2 and 1.

\section{Feedback Accuracy and Student Writing Clarity}
\label{sec:student-clarity}

\begin{figure}[t]
    \centering
    \small
    \begin{tabular}{c|l}
    1 & Relation between mass , PE and KE. \\
    2 & The relationship is related to mass.  \\
    3 & The law of energy, directly related to mass.  \\
    4 & To the law of conservation of energy.  \\
    5 & I am going to explain the science behind why your current roller coaster \\
      & design will be exciting and get to the end of the ride without stopping.  \\
    6 & So we go.  \\
    7 & When I inc I'm going to show you the height of my initial drop and how \\
      & that relates to PE at the top and KE at the bottom.  \\
    8 & rease the initial drop height, the amount of PE and KE conversely, \\
      & when I decrease the initial drop height, the amount of PE KE For example. \\
    9 & As the car traveled down the hill, the PE and the KE transfer the energy \\
      & between each other never going over the initial amount of energy and the \\
      & total energy. \\
   10 & If the kinetic energy goes up, the energy goes down and if the potential energy \\
      & goes up, the kinetic energy will go down.
    \end{tabular}
    \caption{A student essay with very mixed writing quality.}
    \label{fig:sample-essay-mixed-quality}
\end{figure}

The preceding section observed that the main ideas in the roller coaster curriculum differ from one another in regard to the expressive range afforded by the ideas: some ideas are more distinctive than others. This is reflected in closeness in vectors space of pairs of vectors from distinct main ideas (Table~\ref{tab:confusable-main-ideas}). Here we examine a second important factor, one that we later relate to the opportunity for students to reflect on their expressive skills: student writing clarity.  On the one hand, clarity of student writing is fairly obvious to adult readers, as illustrated in the sample essay shown in Fig.~\ref{fig:sample-essay-mixed-quality}.  Given its punctuation, the essay is processed as having ten sentences. Sentences 1-4 and 6 are not full sentences, and 5 merely states the purpose of the essay. The final sentences are much more complete, and fairly clear. (It seems possible that the student mixed up material across sentences 7 through 9, perhaps through faulty cut-and-paste steps.) On the other hand, clarity of writing results from many factors, some of which would be difficult to measure. Instead of attempting to directly measure writing clarity, we assess differences across student essays using cosine similarity of student phrases to main ideas as a probe.  

% Probe 1: average number of MIs > $t$ per content unit
%We binned the 159 essays from GT1 and GT2 into a validation set of 117 and a test set of 42, then the validation set into essays with at most one PyrEval error (N=58; High Acc.), two errors (\textcolor{red}{It does not add up to 159 because there are divided into validation and test set so not all the essays are used here} N=45; Mid Acc.), or more errors (N=14; Low Acc.). 
We randomly selected 117 essays from the 159 for the data exploration reported here, then used the remaining 42 essays to test consistency of our observations. We binned essays into those with at most one PyrEval error (N=58; High Acc.), two errors N=45; Mid Acc.), or more errors (N=14; Low Acc.). We calculated the average number of main ideas per clause whose cosine similarity exceeded $t$, for all essays in a given accuracy bin. clauses in high and mid accuracy essays matched an average of 1.63 main ideas (sd=0.84), while clauses in low accuracy essays matched 1.73 main ideas on average (sd=0.78).  When a given clause is a candidate match for up to 3 ($topk$) content units, it will appear in that many hypergraph nodes, and the algorithm is more likely to pick the least ideal match.

\begin{figure}[t]
    \centering
    \begin{enumerate}
        \item an object has the more PE it will have at the top of the drop and the more total energy (low clarity)
        \item the stored energy will turn into kinetic energy because of the gravity (low clarity)
        \item but, since the law of Conservation of Energy states that energy can not be created or destroyed, the PE, does not just disappear (high clarity)
        \item because, based on the data that was collected, the hill height has to be smaller than the initial drop height (high clarity)
    \end{enumerate}  
        \setlength{\tabcolsep}{10pt}
    \begin{tabular}{l|c|c|c|c|c|c}
    \multicolumn{1}{c}{ID} & 1 & 2 & 3 & 4 & 5 & 6\\\hline
        \multicolumn{7}{c}{Low Clarity Examples} \\\hline
        1 & 0.63  & -  & 0.52   & 0.51  &  - & 0.57 \\
        2 &  - & 0.58  &  -  &  0.53 & -  & 0.52 \\\hline
        \multicolumn{7}{c}{High Clarity Examples} \\\hline
        3 &  - & -  &  - & 0.71  &  - &  -\\
        4 &  - & -  &  - &  - & -  &  0.69\\
    \end{tabular}

    \caption{Clauses with low versus high clarity, and main ideas they are similar to.}
    \label{fig:low-clarity-versus-high-clarity}
\end{figure}

\begin{table}[b!]
    \centering
    \small
    \begin{tabular}{l|r|r}
    \multicolumn{1}{c|}{Pair} &
        \multicolumn{1}{c|}{Count} &
            \multicolumn{1}{c}{Avg. Sim.} \\\hline
    1-5     &  1,152 &  0.53 \\  
    1-2     &    986 &  0.59 \\
    2-6     &    802 &  0.48 \\
    3-6     &    534 &  0.54 \\
    1-6     &    532 &  0.44 \\
    2-5     &    472 &  0.40 \\
    2-3     &    214 &  0.39 \\
    5-6     &    170 &  0.27 \\
    1-3     &    158 &  0.38 \\
    4-6     &     69 &  0.40 \\
    2-4     &     57 &  0.30 \\
    3-5     &     25 &  0.16 \\
    3-4     &     16 &  0.38 \\
    1-4     &      6 &  0.21 \\
    4-5     &      5 &  0.06 \\
    \end{tabular}
    \caption{Confusability of all pairs of main ideas, as average cosine similarity.}
    \label{tab:confusable-main-ideas}
\end{table}

The examples in Fig.~\ref{fig:low-clarity-versus-high-clarity} illustrate the same point about the impact on accuracy if a clause has above threshold similarity to more than one main idea. The top of the figure shows two phrases that are poorly written, and that have cosine similarities above $t$ for multiple ideas. The lower half of the figure shows two well articulated statements, where each is above threshold similarity to exactly one main idea, and also where the cosine similarity is much higher than 0.50. Table~\ref{tab:confusable-main-ideas} shows how often each pair of main ideas is a candidate match for multiple clauses in the set of 150 essays. Pairs of main ideas are shown in descending order of the number of clauses that have a similarity to both above the threshold $t$. The third column, the average pairwise cosine similarity of the vectors in the CUs for each pair, was discussed above. Main ideas 1 and 5 are the most ``confusable'' for PyrEval, having a count of 1,152 clauses that have above similarity $\geq t$ to both. The confusability is also reflected in the high average cosine similarity of their CU vectors. Count and Avg. Sim have a Pearson correlation of 0.78.

\begin{figure}[t!]
    \centering
    \includegraphics[scale=0.75]{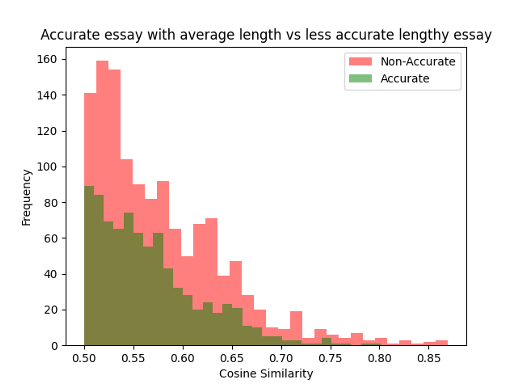}
    \caption{Cosine similarity distributions of clauses in the full assessment hypergraph for an accurate short essay, and a long inaccurate essay.}
    \label{fig:acc-short-vs-inacc-long}
\end{figure}

% Probe 2: comparison of short accurate essay versus long inacc essay (Example 3)
Fig.~\ref{fig:acc-short-vs-inacc-long} plots the cosine similarity (x-axis) by number of clause-main idea pairs at that cosine similarity (y-axis) in an accurately assessed essay of average length versus an inaccurately assessed essay that was quite lengthy.  The accurate essay (darker bars) has a lower count of clauses overall, but more importantly, very few that have a cosine similarity of 0.70 and above.  In contrast, the inaccurate essay has about ten times as many at that cosine similarity and above, which increases the chances that the node selected by the MIS algorithm as a clause matching a main idea would not be one that a human would select.

% Probe 3: main idea 1 versus main idea 6

\begin{figure}[t!]
    \centering
    \includegraphics[scale=0.69]{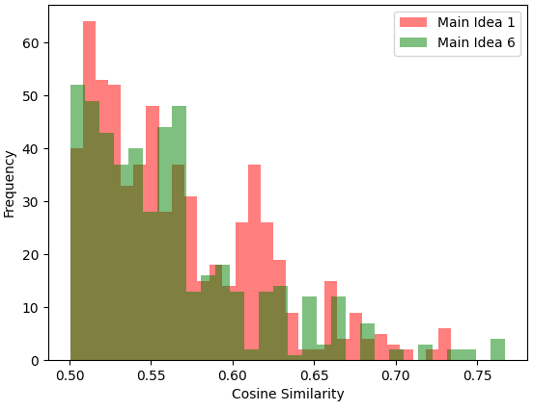}
    \caption{Cosine similarities of clauses in the val. set (N=117) to main idea 6 (high accuracy) vs. main idea 1 (low accuracy). The test set plot looks the same.}
    \label{fig:mainidea6-vs-mainidea1}
\end{figure}

\begin{figure}
    \centering
    \includegraphics[scale=0.5]{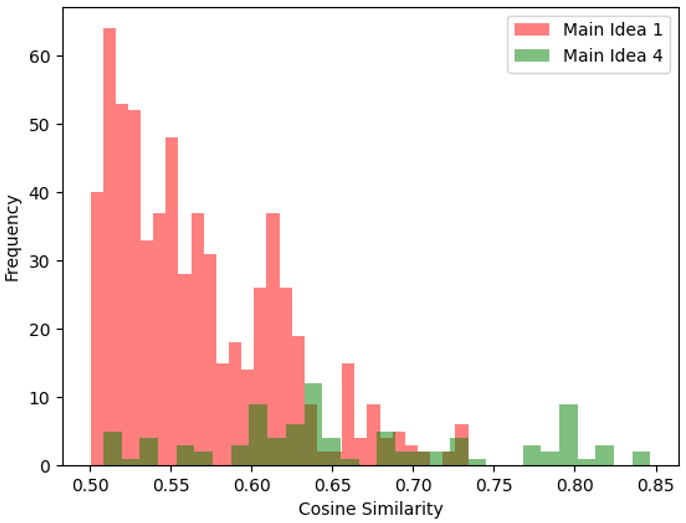}
    \caption{Cosine similarities of clauses in the val. set (N=117) to main idea 4 (high accuracy) versus main idea 1 (low accuracy). The test set plot looks the same.}
    \label{fig:mainidea4-vs-mainidea1}
\end{figure}

Both Figs.~\ref{fig:mainidea6-vs-mainidea1} and \ref{fig:mainidea4-vs-mainidea1} show the number of clauses in the 117 essay subset that have cosine similarities above $t$ for a high accuracy main idea (6 or 4) and main idea 1, which has low overall accuracy. 
Fig.~\ref{fig:mainidea6-vs-mainidea1} for main idea 6 versus main idea 1 % \textcolor{red}{only some of the 159 essays were chosen for the experiments. They are 117 set of randomly chosen essays} for the 117 validation set essays. 
shows that main idea 1 (orange bars) has many candidate matches with cosine similarities of 0.66 and above, whereas main idea 6 has very few.  When there are lower counts at the higher cosine values, it is more likely that these clauses are ``true'' matches to the relevant idea, e.g., main idea 6. As discussed in regard to the preceding figure, when there are higher counts at the high cosine values, it is more likely that some of those clauses are not true matches, e.g., main idea 1, where ``true'' means a clause that a human would identify as a match.
%when there are many candidate matches with a very high cosine similarity is that this increases the chances that the MIS algorithm will select a clause-main idea pair that would not be one that a human would select.

In general, plots of the cosine similarity distributions of student clauses to main ideas have a pattern where the high accuracy essays have distributions that extend to relatively higher cosine values of 0.70 and above but with low counts. For the lower range of cosine similaritie values between 0.50 and 0.70, there are typically many clauses, as in Figs.~\ref{fig:acc-short-vs-inacc-long}-~\ref{fig:mainidea6-vs-mainidea1}. In contrast, the distribution of cosine similarities to main idea 4, the law of conservation of energy, is quite distinctive. Fig.~\ref{fig:mainidea4-vs-mainidea1} plots cosine similarities of all clauses from student essays
%\textcolor{red}{This plot is the same as plot 5. It is not divided by essay accuracy. it just shows pairwise cos sim on a high accuracy main idea vs a lower accuracy main idea} 
to main idea 4 versus main idea 1. For main idea 4, there are far fewer student clauses that have similarities between 0.50 and 0.70. This is further evidence of the distinctiveness of main idea 4, i.e., statements of the LCE.

% Probe 4: high/mid/low main idea 6

% \begin{figure}
%     \centering
%     \includegraphics{figs/mainidea6-high-mid-low.png}
%     \caption{Cosine similarities (x-axis) by count of clauses at that similarity (y-axis) that are candidates to match main idea 6, from essays with high, mid and low PyrEval accuracy.}
%     \label{fig:mainidea6-high-mid-low}
% \end{figure}
% % add correlation of accuracy with length? i.e., average length of high acc, avg length of mid acc, average length of low acc in word tokens

% Fig.~\ref{fig:mainidea6-high-mid-low} is another plot of cosine similarity by count of clauses. It represents similarity distributions to the content unit for main idea 6, drawing from all essays that have high, mid, or low PyrEval accuracy \textcolor{red}{on main idea 6?}.

\section{Discussion}
\label{sec:discussion}

A semantic vector identifies a point in $N$-dimensional space, where the value of $N$ is a hyperparameter for the user to specify during training.\footnote{The WTMF vectors used here are 100D, and the BERT vectors are 768D.}  Similarity of meaning is defined as nearness in the $N$-dimensional space, using cosine as a metric. In PyrEval, a content unit of weight 5 defines 5 nearby points, which essentially expands the meaning of a main idea from a single point to a small region. This helps make PyrEval more robust at identifying the meanings of students' statements with main ideas. While semantic distinctiveness versus vagueness cannot be directly represented in vector space, %, any more than compositionality of meaning, that is, that a sentence is composed of phrases, which are composed of words. 
we presented indirect evidence of the inherent distinctiveness of certain ideas, along with vagueness in students' statements, through examinations of distributions of cosine similarities of clauses from students' essays to content units representing main ideas.  %Thus we see in the contrast between Fig.~\ref{fig:mainidea6-vs-mainidea1} and Fig.~\ref{fig:mainidea4-vs-mainidea1} evidence of the greater distinctiveness of the LCE (idea 4), as well as its low rank in most pairs in Table~\ref{tab:confusable-main-ideas}.  
This suggests that future work on writing assessment could benefit by incorporating methods for ranking the specificity of statements~\cite{huang-etal-2021-definition}.

% When we consider individual statements within an essay, distribution of a statement's cosine similarities across all six main ideas is another way to identify distinctiveness of statements, as illustrated in Fig.~\ref{fig:low-clarity-versus-high-clarity}.  There we saw two statements from student essays that are vague and less well-articulated. This is reflected in the observation that both have cosine similarities somewhat above threshold with three of the main ideas. The two other statements shown there are very clear and specific, and have high similarity to exactly one main idea.  

In our discussions with teachers and observations of students in the classroom, we found that PyrEval provides a second level of feedback that was not part of our original aim.  Our original goal was to provide sufficiently accurate feedback to help students revise their essays by adding ideas that were missing from their original submissions.  We found that when PyrEval is inaccurate, this can result from lack of clarity in student writing. This in turn can help students reflect on how to reduce vagueness and improve clarity in their writing. %, as in the first two statements of Fig.~\ref{fig:low-clarity-versus-high-clarity}, or that are incomplete, as in the first half of Fig.\ref{fig:sample-essay-mixed-quality}.

\section{Conclusion}

Through error analysis of a software tool that provides formative feedback on students' science explanation essays, we have presented two perspectives on distinctiveness of ideas.  One perspective is that when a curriculum aims for students to understand certain science ideas, the ideas themselves have different degrees of distinctiveness.  A science idea that becomes formalized as a natural law, such as the law of conservation of energy, is often associated with a relatively formulaic statement that relies on specific vocabulary. Other ideas, such as how potential energy of an entity relates to its kinetic energy, can be expressed in a wide variety of ways, so long as the trade-off between the two is stated. A second perspective is that students' statements of an idea can be more or less clearly articulated.  Both factors affect the accuracy of a software tool we employed to provide formative feedback on students' essays. This suggests that use of automated AI methods to support classroom instruction can lead to more fine-grained insights into how to articulate science ideas for students, and how to help students learn to articulate their own ideas.

\bibliographystyle{splncs04}
\bibliography{aied2024}

\end{document}